\documentclass[runningheads]{llncs}
\usepackage[T1]{fontenc}
\usepackage[colorinlistoftodos]{todonotes}

\usepackage{graphicx} 
\usepackage[graphicx]{realboxes} 
 \usepackage{booktabs} 
\usepackage{xcolor}

\usepackage{amsmath}
\usepackage{amsfonts}
\usepackage{amssymb}
\usepackage{adjustbox}

\usepackage{enumitem}
\usepackage{subfig}
\definecolor{segment1}{HTML}{FF7F50} 
\definecolor{segment2}{HTML}{1E90FF} 
\definecolor{segment3}{HTML}{32CD32} 
\definecolor{segment4}{HTML}{FF1493} 

\begin{document}
\title{Dynamic Interpretability for Model Comparison via Decision Rules}
\titlerunning{Dynamic Interpretability for Model Comparison via Decision Rules}
%

\author{Adam Rida\inst{1} \and
Marie-Jeanne Lesot\inst{1} \and
Xavier Renard\inst{2} \and
Christophe Marsala\inst{1}
}
%
\authorrunning{A. Rida, M.-J. Lesot, X. Renard, C. Marsala}
%

\institute{Sorbonne Université, CNRS, LIP6\\ \email{\{adam.rida, marie-jeanne.lesot, christophe.marsala\}@lip6.fr}\\
\and
AXA, Paris, France \email{xavier.renard@axa.com}\\
} 
\maketitle
\begin{abstract}
Explainable AI (XAI) methods have mostly been built to investigate and shed light on single machine learning models and are not designed to capture and explain differences between multiple models effectively. This paper addresses the challenge of understanding and explaining differences between machine learning models, which is crucial for model selection, monitoring and lifecycle management in real-world applications. We propose DeltaXplainer, a model-agnostic method for generating rule-based explanations describing the differences between two binary classifiers.
To assess the effectiveness of DeltaXplainer, we conduct experiments on synthetic and real-world datasets, covering various model comparison scenarios involving different types of concept drift.
\end{abstract}
\section{Introduction}
Machine learning (ML) models play a crucial role in solving critical real-world problems, such as fraud detection or asset pricing, across various domains. These models are not static objects: the context in which they operate is often subject to change, which can have a direct impact on their performances.
Phenomena like adding new instances gathered over time to populate the training set or the presence of concept drift, when the data distribution changes over time~\cite{breck_2019} can lead to a decrease in prediction quality over time.
To ensure constant or improved performances, a deployed ML model has to be 
updated in response to the evolving context: as with any product, deployed ML models have a lifecycle, from development to update, through monitoring. 

Yet, most of the literature in explainable AI (XAI)  focuses on explaining the behavior of individual ML models.
Now, it has been shown~\cite{kulesza_mental_2012} that ML model updates can disrupt the mental models users construct about these ML models to understand their behaviors when they experience different outcomes. This disruption can affect users' decision-making and eventually weaken their trust in the ML model.  
Thus having \textit{differential explanations} to explain what has changed between successive versions of an ML model could help users, among other benefits, to update their mental models.
In this paper, we address this problem, which we call \textit{differential explainability}.

To characterize the differences between two ML models, we propose a framework called $\Delta$-modelling.
The idea consists of a decomposition of the differences between two black-box ML models into a set of interpretable and complementary models, each covering subspaces of the model domain where the two ML models have different behavior.
We instantiate this framework with a method called DeltaXplainer that outputs rule-based descriptions of the areas where prediction changes between the two models. 
We experiment DeltaXplainer on several real-world and synthetic datasets 
to show its performances to capture the models' differences in an interpretable way.

The paper is organized as follows. After a brief review of related works in Section~\ref{sec:edla}, Section~\ref{sec:probSetting} specifies the considered problem setting and the general principle of the proposed approach. Section~\ref{sec:prop} describes its instantiation, DeltaXplainer, in more detail. Section~\ref{sec:expe} presents the experiments conducted on real-world and synthetic data, to assess the accuracy and interpretability potential of DeltaXplainer's generated explanations.
\section{Related Works}
\label{sec:edla}
Training a machine learning model on a given dataset can lead to numerous models with similar performances but differing underlying patterns, a phenomenon called~\textit{Rashomon effect}~\cite{leventi_2022} or model multiplicity~\cite{marx_predictive_multiplicity_2020,black_multiplicity_2022}.
The simultaneous existence of these models presents risks, especially in terms of fairness and robustness, as well as for the transparency and trust users have in the selected model; issues that are not addressed by classical aggregated performance measures~\cite{black_multiplicity_2022}.

It has been proposed to detect a change in a model statistically~\cite{bu_2018,geng_2019,harel_2014}, but change detection does not give any insight into which subpopulations of the domain have seen a change in model behavior, in particular prediction change. Severity measures for predictive multiplicity have been proposed~\cite{marx_predictive_multiplicity_2020}, with the same limitation. Addressing the need for explanations regarding predictive multiplicity or discrepancies among ML models' behaviors has gathered attention in recent research. To overcome the limitations of traditional XAI techniques like LIME~\cite{ribeiro_2016_lime}, Anchor-LIME~\cite{ribiero_2018}, or SHAP~\cite{lundberg_2017}, which primarily focus on explaining the behavior of \textit{one} model, several propositions have emerged. These propositions aim to comprehensively understand and compare differences in ML models' behaviors.

The Discrepancy Interval Generation (DIG)~\cite{renard_understanding_2021} method detects model discrepancies by creating \textit{counterfactual intervals} to explain where the models of interest, trained on the same dataset, disagree.
While this approach provides insights into the differences between models, it does not explain the actual subspace of model discrepancies nor provides a clear identification of the model discrepancies at a global scale.
Nair et al.~\cite{nair_2021} propose a method to compare models by approximating them with interpretable rule sets and grounding to compare explanations.
However, this approach relies on user input to generate explanations that are based on a comparison between the rule-based surrogates built for each of the two models to compare. In contrast, we use rules as our final explanations, built to predict their disagreement.

These  methods  address the concept of predictive multiplicity or model discrepancies but do not take into account the dynamic nature of the task and the underlying data distribution over time. For instance, they do not consider scenarios where the dataset and data distribution evolve due to concept drift. Recent research proposes  to tackle this aspect by observing the evolution of feature importance in models over time~\cite{muschalik_2022} or identifying the key features contributing to the observed differences~\cite{wang_2022}.

Recent research has focused on explaining dataset drift~\cite{hinder_2023}. The authors propose training a model to classify instances from the dataset into pre-drift and post-drift periods. By generating explanations for this model, they aim to pinpoint the specific areas where the drift occurred. However, this approach does not specifically address the differences in behaviors between trained models induced by concept drift. 

The aim of this paper is to propose a method that generates explanations for the discrepancies in behaviors between two models. The second one is trained on the same task as the first one but using a dataset that has undergone concept drift. Our approach aims to describe and help users understand the specific subspaces or sub-populations where the two models exhibit divergent behaviors. Unlike previous works that primarily focus on feature importance, the DeltaXplainer method we propose provides insights into the variations of behavior between the models.
\section{Problem Setting}
\label{sec:probSetting}
This section sets the scope of the problem of generating explanations with the aim to describe machine learning model differences in a dynamic setting. It also discusses and motivates the choice of decision rules as explanations.
\subsection{$\Delta$-modeling and Differential Explainability}
\begin{figure}[t]
\centering
\includegraphics[width=0.75\textwidth]{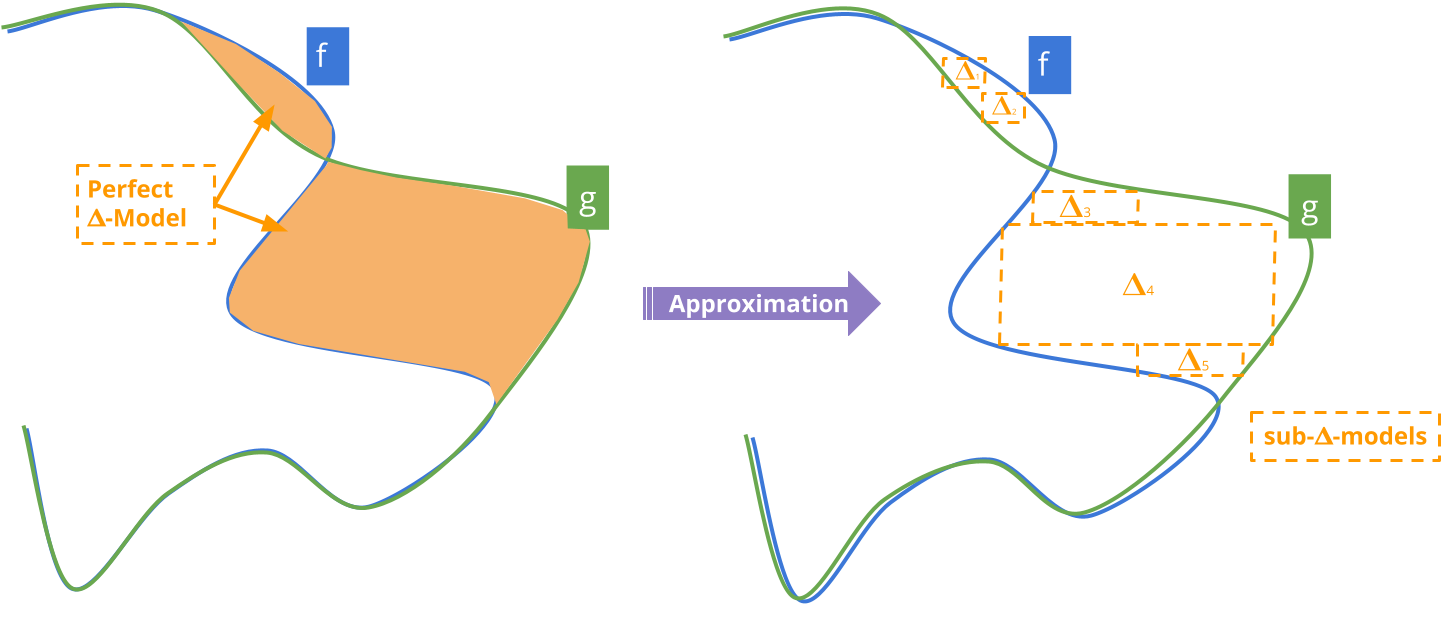}
\caption{$\Delta$-modelling principle: training a model to predict the differences. (Left) Perfect model, which is only  reachable in theory; (right) proposed approximation through decision rules.
} \label{fig_schema_perfect_delta_model}
\end{figure}
We consider two binary black-box classifiers $f$ and $g$ with their respective training data sets $(X_f,y_f)$ and $(X_g,y_g)$. We assume the descriptive features are numerical 
and 
both classifiers are defined on $\Omega \mapsto \{0,1\}$ with $\Omega \subset \mathbb R^d$ where $d$ denotes the number of features and $\Omega$ is known and bounded (i.e. the minimum and maximum possible values are known for each feature).

We describe a dynamic setting as a scenario involving the evolution of either the data (e.g. drift) or the models (e.g. model update). Therefore
we propose to formalize the differential explanation generation issue as the study of a third model trained to predict where the models disagree, as illustrated on Fig.~\ref{fig_schema_perfect_delta_model}. 
We define a $\Delta$-model, as a binary classifier also defined on $\Omega \mapsto \{0,1\}$, that predicts, for any $x \in \Omega$: 
\begin{equation}
\label{equation_delta_model}
\begin{cases}
1 & \text{if } f(x) \neq g(x), \\
0 & \text{otherwise}
\end{cases} 
\end{equation}
Such a $\Delta$-model can then be used to identify the regions of the input space where the predictions of $f$ and $g$ disagree. Note that access to the inner structure of $f$ and $g$ is not needed: this makes $\Delta$-modelling model-agnostic.

The $\Delta$-model then contains information regarding what differs between the two models and, similarly to~\cite{hinder_2023}, XAI can be leveraged to  extract insights from this model. 
As such, the main challenges consist of modeling properly the differences and simultaneously being able to generate explanations of these differences, which may raise the question of the interpretability-accuracy trade-off. We propose to train an approximation of  the true model differences, as illustrated in the right part of Fig.~\ref{fig_schema_perfect_delta_model},  using a decision rule approach.  
\subsection{Rules as Differential Explanations}
We define a rule-based explanation to describe differences between models as a set of differential rules, $\left \{ r_i \right\}, i = 1,...,p$.
This explanation is designed to describe exhaustively the areas of differences between models. 
Each differential rule $r$ is a set of conjunctions delimiting a subspace of $\Omega$:
\begin{equation}
r= \bigwedge_{j=1}^{d}x_j\in [a_j,b_j] 
\end{equation}
with $d$ the total number of features $x_j$ the $j$-th feature, $[a_j,b_j]$ its associated range of values. Obviously, not all features appear in all rules: for irrelevant features, the interval equals the total feature range, which is assumed to be known. The rule length is defined as the number of relevant features. 

We choose rule-based explanations for characterizing the differences between models due to their intuitive and interpretable nature. Each rule represents a specific region of the input space where the predictions of $f$ and $g$ disagree, providing insights into the underlying factors causing the disagreement. In comparison, global XAI methods or feature importance-based explanations are not designed to capture specific regions in a feature space and usually do not provide information on feature interactions

Still, most rule extraction methods, such as deriving them from a large decision tree or using Rulefit~\cite{friedman_2008}, do not guarantee the interpretability of the rule set. Indeed, they may generate a high number of rules with high length, contradicting the interpretability criteria of sparsity and simplicity, e.g. established in~\cite{miller_explanation_2018}. They may also generate overlapping rules, with partial redundancies, as is for instance the case of  Sequential Covering~\cite{cohen_1995} or rules derived from Random Forests, again harming interpretability and requiring potentially expensive post-processing. On the other hand, oversimplifying rules might decay their predictive performances and hence provide misleading explanations.

To quantify the appropriateness of a set of rules $\mathcal R = \left \{ r_i \right\}, i = 1,...,p$, we propose to use the following criteria as proxy for interpretability: 
\label{sub:mesures}
\begin{itemize}
  \item Minimize the number of rules, denoted \textbf{\#r}
  \item Minimize the length of each rule, denoted \textbf{\#l}
  \item Maximize the coverage of each rule, denoted \textbf{cov}
  \item Minimize the overlap between rules
\end{itemize}
The number of rules corresponds to the $p$ parameter of $\mathcal R$. Coverage is defined as the proportion of data points that it covers, i.e. that trigger it, compared to the total number of data points, it corresponds to its support. The length and coverage criteria are defined for individual rules, they are averaged across the rule set. Other aggregation operators may be considered, depending on the user preferences: a user may be interested in rules with similar coverage or, on the contrary, with rules with a high variance of coverage. 
Note that these four criteria can be correlated: in cases without overlap between rules, increasing coverage will mechanically reduce the number of rules. 

Additionally, we use the following criteria as proxy for the rule set fidelity wrt the models whose difference must be explained, i.e. wrt the classification task defined in Eq.~(\ref{equation_delta_model}):
\begin{itemize}
  \item Maximize the accuracy
  \item Maximize the precision of the disagreement class
  \item Maximize the recall of the disagreement class
\end{itemize}
\begin{figure}[t]
\centering
\includegraphics[width=0.8\textwidth]{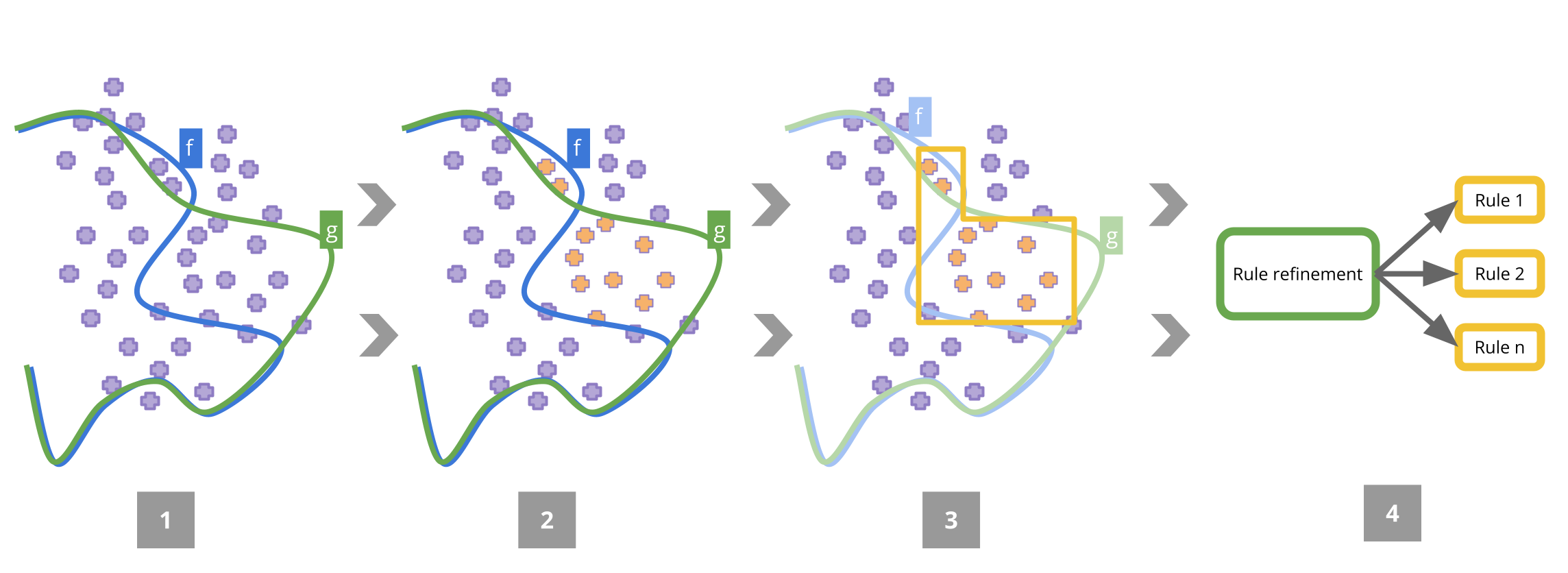}
\caption{Four steps of the DeltaXplainer algorithm} \label{fig_deltaxplainer}
\end{figure}
\section{Proposed Method:  DeltaXplainer}
\label{sec:prop}
This section describes DeltaXplainer, the method we propose to generate interpretable rule-based explanations to describe differences between models. 
As illustrated in Fig.~\ref{fig_deltaxplainer}, DeltaXplainer is a 4 step procedure taking as input $f$, $g$ and their training data  $(X_f,y_f)$ and $(X_g,y_g)$. 
\subsubsection{Steps 1 and 2: Constructing the training data $(X_{\Delta}, y_{\Delta})$}
DeltaXplainer builds a surrogate model  trained on samples labelled according to the agreement of $f$ and $g$. As we consider a dynamic setting, we assume that areas of interest lie within the distribution of the data. Therefore, in Step~1, the training set is defined as $X_{\Delta} = X_f \cup X_g$. In Step~2, their labels $y_{\Delta}$ for $X_{\Delta}$ are set using $f$ and~$g$ according  to Eq.~\ref{equation_delta_model}.
\subsubsection{Step 3: Fitting a Decision Tree}
We propose to consider as a surrogate model a decision tree, due to the possibility to have some control over its interpretability: its  hyperparameters have a significant impact on its structure, which steers the interpretability of the explanations that can then be extracted from it. For example, increasing the minimum number of samples per leaf leads to areas with higher coverage and restraining the maximum depth generates fewer branches and hence fewer rules.
Another advantage of using a binary decision tree is to prevent rule overlap, as a sample cannot be covered by multiple rules.

In order to ease the choice of a good set of hyperparameters to reach an interpretability-accuracy trade-off acceptable for the user and avoid over-cons\-training the model that would limit its performances, we propose to restrict the model parametrization to the minimum number of samples per leaf, without constraining the maximum depth of the tree. As there is no overlap in a tree, both parameters are correlated, preventing uncontrolled growth of the tree.
\subsubsection{Step 4: Rule Extraction}
The final step consists of translating the tree into a set of rules, associating one rule to each branch, and keeping only the ones predicting class~1 (disagreement class): each of them characterizes an area where $f$ and $g$ differ in prediction. In addition, the successive tests performed in each tree node are rewritten to improve interpretability: they are grouped depending on the feature they apply to, to make explicit the value interval they define. For instance, this post-processing step transforms  “if {$(A > 10)$} and {$(A \geq 30)$} and {$(A \leq 50)$}” to “if {$A \in [30, 50]$}". This refinement chooses to omit the hierarchical structure of the rule set and to favor a feature-based view, decreasing the length of the rule expression by decreasing the number of conjunction it contains.  
\section{Experiments}
\label{sec:expe}
This section describes the experiments conducted to assess DeltaXplainer performance, in particular its ability to 
(1) capture the differences in prediction between models and (2) provide  explanations of the areas with such differences. The experiments are performed
in dynamic settings, considering different forms of data drifts, 
on synthetic and real datasets. After a description of the experimental protocol, the results are discussed in turn  qualitatively and quantitatively.
\subsection{Experimental Protocol}
\subsubsection{Datasets \& Model Type}
The model retraining scenarios we consider are built upon datasets from the concept drift literature: AGRAWAL~\cite{agrawal_1993} as synthetic data, COVER TYPE~\cite{blackard_1998} and ELEC2~\cite{harries_1999} as real-world data.

We consider these datasets as a stream of data following their time indexation or the order they have been generated for synthetic data. 
We then sub-sample a sub-stream of 10000 samples. The data, restricted to its numerical features, is normalized before applying the following modifications, and the inverse standardization is applied at the end of the overall process to generate explanations in the original domain. 
Then, we apply the protocol described in~\cite{hinder_2023} which consists in splitting the sub-stream at a random point between 1/3 and 2/3 of the data set.
A perturbation is applied on the second split to simulate a drift, as  described in the next paragraph. 

The first split is denoted $X_f$ and the union of the first and second splits $X_g$. Model $g$ can then be seen as a retrained version of model $f$. Models $f$ and $g$ are instantiated with random forests of 100 estimators. We split $X_f\cup X_g$ into a train and test set (70\%/30\%). 
\subsubsection{Perturbations and Model Comparison Scenarios}
We assess DeltaXplainer in three dynamic contexts elaborated from~\cite{hinder_2023} by changing the type of perturbation applied to the second split of the data: Gaussian noise, permutation and shift respectively denoted S1, S2 and S3.

\textbf{S1} introduces small and subtle changes between models by adding Gaussian noise to a random number of features. It simulates a context where models are exposed to minor perturbations such as noise in the measurements or sampling. 
\textbf{S2} permutes feature values,  simulating a situation where models differ sparsely, with differences challenging to capture. 
\textbf{S3} adds a fixed shift (adding 2 to the feature values in our case) to a random number of features, ranging from 2 to the maximum number of features. It simulates a context where models are exposed to a significant distribution change, with an abrupt drift.
\subsubsection{DeltaXplainer Configuration} 
 The $\Delta$-model is trained on the training set built from $X_f\cup X_g$.
 As a baseline to assess its fidelity to the differences in prediction between $f$ and $g$, the decision tree is set with no constraints to grow.
  We vary the considered tree hyperparameter, the minimum number of samples per leaf, from the smallest possible value, setting to 1, to various proportions of the total number of samples, considering the ratios 0.1\%, 1\%, 2.5\%, 10\%, and 25\%. This ensures that no rule  has a coverage lower than this hyperparameter, which allows controlling  directly one of the interpretability measures.
\subsection{Qualitative Results}
We illustrate the output of DeltaXplainer with explanations generated on the AGRAWAL dataset under scenario S3 with a 1\% minimum sample per leaf. This synthetic dataset describes a loan approval use-case where features are the applicant's characteristics. The shift perturbation is here applied on the age, salary and education level.

The explanation of the differences in prediction between models $f$ and $g$, i.e. before and after the drift, consists in the following 4 rules:
\begin{itemize}
\item \textbf{R1:} \textcolor[HTML]{1E90FF}{(salary $>$ 170824.88)} and   \textcolor[HTML]{ED7D31}{(education level $>$ 9.17)} and   \textcolor[HTML]{32CD32}{(age $\leq$ 129.62)} (199 samples, coverage 2\%))
\item \textbf{R2:}  \textcolor[HTML]{1E90FF}{(salary $>$ 170824.88)}  and  \textcolor[HTML]{ED7D31}{(education level $\leq$ 6.17)}  and  \textcolor[HTML]{32CD32}{(age $>$ 109.62)} (182 samples, coverage 1.8\%)
\item \textbf{R3:} \textcolor[HTML]{1E90FF}{(salary $>$ 170824.88)}  and  \textcolor[HTML]{ED7D31}{(education level $\in [7.17, 9.17]$)}  and  \textcolor[HTML]{32CD32}{({age~$\leq$~ 109.62})} (158 samples, coverage 1.6\%)
\item \textbf{R4:}  \textcolor[HTML]{1E90FF}{(salary $>$ 170824.88)}  and  \textcolor[HTML]{ED7D31}{(education level $\in$ [6.17, 7.17])} and \textcolor[HTML]{32CD32}{(age~$>$~129.62)} (101 samples, coverage 1\%)
\end{itemize}
These 4 rules represent where the model predictions differ and the support of each rule (how many samples are affected). Each of these explanations can be used as a local explanation, for instance, to explain why an instance has different predictions across models.
In practice, this explanation can be used to investigate the sources of the differences in prediction. We can observe that differences occur for extreme values (as the shift is abrupt on the age, education level and salary) showing that the differences between the two models are actually from areas where the drift happened. The explanation can also be used to investigate whether the difference in performances of the two models is caused by the drift.
\subsection{Quantitative Results}
For the quantitative assessment, we consider the two categories of metrics as introduced in Section~\ref{sub:mesures}: regarding the fidelity of the $\Delta$-model to capture the differences between $f$ and $g$, we consider the recall, accuracy and precision of the global $\Delta$-model on the test set. Note that, depending on the scenario, the two classes (agreement/disagreement) may be unbalanced: for S1 and S2 as the changes are small, only a few samples are actually different. For this reason, accuracy is mainly relevant for S3 (balanced data set). 

Regarding the interpretability of the generated set of rules, we consider the number of rules, their average length and their average coverage. Because we are using these metrics as a proxy for interpretability, we are interested in measuring them across all $\Omega$. This is particularly true for the coverage, we want to approximate the area covered by each rule. Therefore, for these metrics, we use the whole data set, training and testing. 

All measures are computed only in the case where at least one rule characterizing the disagreement area can be extracted, depending on the value set to the hyperparameter. 
Table~\ref{tab_result_elec}  provides the average and standard deviation on 10 runs for the ELEC2 data set, the appendix shows the results for the other data sets, that globally lead to the same analyses and conclusions.

It shows that coverage is low. We study in this experiment how it evolves and correlates to the minimum sample hyperparameter. The idea is to observe whether increasing the minimum sample per leaf indeed increases interpretability (through our proxy metrics). We also want to observe how big the drop in performance is.
\begin{table}[t]
\caption{Experimental results for ELEC2, avg and std over 10 runs}\label{tab_result_elec}
\resizebox{\textwidth}{!}{%
\begin{tabular}{|c|c|c|c|c|c|c|c|}
\hline

Scenario & Min samples & Acc &Prec & Rec &\textbf{\#r} & Mean \textbf{\#l} & Mean \textbf{cov} \\
\hline

 &1 sample &0.93 $\pm$ 0.03 &0.39 $\pm$ 0.14 &0.39 $\pm$ 0.07 & 61.5 $\pm$ 10.32 &7.64 $\pm$ 0.36 &0.1\% $\pm$ 0.05 \\
 & 0.1 \% &0.94 $\pm$ 0.01 & 0.44 $\pm$ 0.1 & \underline{\textbf{0.4}} $\pm$ 0.08 & 31.9 $\pm$ 6.1 & 6.91 $\pm$ 0.4 &0.15\% $\pm$ 0.02 \\
S1 & 1 \% &\underline{\textbf{0.95}} $\pm$ 0.01 &\underline{\textbf{0.52}} $\pm$ 0.12 &0.25 $\pm$ 0.15 & 2.0 $\pm$ 0.82 &4.55 $\pm$ 0.73 &1.2\% $\pm$ 0.26 \\
& 2.5 \% &0.38 $\pm$ 0.49 &0.21 $\pm$ 0.27 &0.16 $\pm$ 0.22 & 0.5 $\pm$ 0.71 & 1.2 $\pm$ 1.62 &\underline{\textbf{1.4\%}} $\pm$ 1.87 \\
 &10 \% &- &- &- &0.0 &- &- \\
 &25 \% &- &- &- &0.0 &-&- \\
\hline
 &1 sample &0.59 $\pm$ 0.02 &0.55 $\pm$ 0.03 &0.51 $\pm$ 0.03 &549.6 $\pm$ 57.91 &8.69 $\pm$ 0.85 &0.08\% $\pm$ 7e-03 \\
 & 0.1 \% & 0.6 $\pm$ 0.02 &0.55 $\pm$ 0.02 &\underline{\textbf{0.54}} $\pm$ 0.02 & 302.6 $\pm$ 28.3 &7.95 $\pm$ 0.71 &0.15\% $\pm$ 0.01 \\
S2 & 1 \% & 0.6 $\pm$ 0.03 &0.57 $\pm$ 0.04 &0.51 $\pm$ 0.05 &30.3 $\pm$ 5.46 &5.46 $\pm$ 0.32 &1.36\% $\pm$ 0.18 \\
 & 2.5 \% &0.62 $\pm$ 0.03 &0.59 $\pm$ 0.03 &0.51 $\pm$ 0.09 &12.1 $\pm$ 2.88 &4.19 $\pm$ 0.27 &3.29\% $\pm$ 0.27 \\
 &10 \% &\underline{\textbf{0.62}} $\pm$ 0.03 & \underline{\textbf{0.6 }}$\pm$ 0.03 &0.47 $\pm$ 0.08 & 2.6 $\pm$ 0.52 &2.23 $\pm$ 0.33 &13.82\% $\pm$ 1.6 \\
 &25 \% & 0.6 $\pm$ 0.03 &0.59 $\pm$ 0.05 &0.45 $\pm$ 0.11 & 1.2 $\pm$ 0.42 & \underline{\textbf{1.1 }}$\pm$ 0.21 &\underline{\textbf{29.83}}\% $\pm$ 4.97 \\
\hline
 &1 sample &0.97 $\pm$ 0.01 &\underline{\textbf{0.86}} $\pm$ 0.05 &0.84 $\pm$ 0.05 &45.0 $\pm$ 7.15 &6.88 $\pm$ 0.55 &0.25\% $\pm$ 0.039 \\
 & 0.1 \% &\underline{\textbf{0.97}} $\pm$ 0.01 &0.86 $\pm$ 0.04 &0.85 $\pm$ 0.05 &34.2 $\pm$ 6.97 &6.17 $\pm$ 0.73 & 0.33\% $\pm$ 0.06 \\
S3 & 1 \% &0.96 $\pm$ 0.01 &0.86 $\pm$ 0.05 &0.82 $\pm$ 0.06 & 7.5 $\pm$ 1.96 & 4.41 $\pm$ 0.7 &1.5\% $\pm$ 0.303 \\
 & 2.5 \% &0.96 $\pm$ 0.01 &0.84 $\pm$ 0.07 &0.81 $\pm$ 0.09 & 3.4 $\pm$ 0.97 &3.51 $\pm$ 0.64 &3.33\% $\pm$ 0.424 \\
 &10 \% &0.83 $\pm$ 0.29 &0.56 $\pm$ 0.21 &\underline{\textbf{0.88}} $\pm$ 0.31 & 0.9 $\pm$ 0.32 & 1.0 $\pm$ 0.47 &\underline{\textbf{16.68\%}} $\pm$ 6.541 \\
 &25 \% &- &- &- &0.0 &- &- \\
\hline

\end{tabular}%
}
\end{table}
\subsubsection{Fidelity} With the baseline instantiation (minimal sample per leaf = 1) we observe that the DeltaXplainer performances to capture differences vary significantly across the 3 considered scenarios. Changes induced by S1 and S2 are the most difficult to capture, which is expected, as these two scenarios trigger sparse and subtle distortions of the decision boundary: areas of differences are small and difficult to capture.
Even if accuracies on S1 are better than on S2, models on S1 struggle significantly more to capture all the correct differences (lower recall).
On the other hand, the changes due to an abrupt shift in the data (S3) are almost perfectly captured. This behavior is also expected, as an abrupt shift in the data triggers a much easier-to-apprehend change in the decision boundary: clear changes in the decision boundaries and grouped affected samples.

We also observe that increased interpretability constraints on the DeltaXplainer (higher support per leaf) decrease almost all performance metrics in the three scenarios. For cases with a value higher than 10\%, we observe a significant drop of performances across the scenarios. The only exception is the recall for S3 with 10\% minimum samples. In cases where no rule has been found, we don't compute evaluation metrics. This can be explained by the fact that precision decreases. This means that the rules cover more samples, increasing the likelihood of capturing points with disagreeing predictions, hence the good recall.
\subsubsection{Interpretability} 
 As desired, controlling the minimum sample per leaf allows for a controlled interpretability of the rules.
 S3 (abrupt shift) obtains overall the best fidelity but is also the most interpretable according to the considered metrics: it has on average the shortest rules 
 and the lowest number of rules generated,  It also has a higher coverage. As already mentioned, the differences in prediction between $f$ and $g$ induced by the abrupt shift of S3 are grouped in the feature space, which is easier to localize and capture for the DeltaXplainer.

 S2 (permutation) leads to the worst interpretability performance. The cause probably lies in the random permutations that generate a lot of local changes that are hard to model: many specific rules are required to capture them.

Interestingly, the number of rules generated with the noise scenario~S1 reaches~0 when the minimum sample per leaf is above 10\%, which happens faster than for the two other scenarios. This can probably be explained by small differences harder to capture for DeltaXplainer. In contrast, permutation (S2) is the only scenario that still generates a rule when this parameter is set to 25\%.
\subsubsection{Conclusion} 
We observe that DeltaXplainer performs differently across the various scenarios S1, S2 and S3.
In situations involving simple drifts, such as abrupt data shifts, DeltaXplainer generates accurate and interpretable explanations. 
In more complex scenarios, such as small and sparse drifts, additional efforts are necessary to achieve satisfactory explanations, with hyperparameters favoring fidelity at the expense of lower interpretability.
\section{Conclusion and Future Works}
In this paper, we introduced and tested a model-agnostic method for generating human-interpretable explanations of model differences in binary classification problems with tabular data in a dynamic setting involving the evolution of the data (e.g., drift).
We proposed a new framework for studying differences between models, called $\Delta$-modelling. We instantiated this framework with DeltaXplainer, an algorithm to generate refined rule-based explanations for finding disagreement areas.
The experiments showed the effectiveness of DeltaXplainer to capture these areas and the influence of the single user-defined parameter on the interpretability-fidelity trade-off.
We illustrated with examples the explanations generated with DeltaXplainer. 

Future works include conducting human-in-the-loop evaluations of generated explanation interpretability, experimenting with other dynamic settings, and using the proposed model at a local scale to explain specific instance prediction changes.


\bibliographystyle{splncs04}
\bibliography{biblio}

\begin{thebibliography}{10}
\providecommand{\url}[1]{\texttt{#1}}
\providecommand{\urlprefix}{URL }
\providecommand{\doi}[1]{https://doi.org/#1}

\bibitem{agrawal_1993}
Agrawal, R., Imielinski, T., Swami, A.: Database mining: a performance perspective. IEEE Transactions on Knowledge and Data Engineering  \textbf{5}(6),  914--925 (1993). \doi{10.1109/69.250074}

\bibitem{black_multiplicity_2022}
Black, E., Raghavan, M., Barocas, S.: Model multiplicity: Opportunities, concerns, and solutions. In: Proceedings of the 2022 ACM Conference on Fairness, Accountability, and Transparency. p. 850–863. FAccT '22, Association for Computing Machinery, New York, NY, USA (2022). \doi{10.1145/3531146.3533149}, \url{https://doi.org/10.1145/3531146.3533149}

\bibitem{blackard_1998}
Blackard, J.: {Covertype}. UCI Machine Learning Repository (1998), {DOI}: https://doi.org/10.24432/C50K5N

\bibitem{breck_2019}
Breck, E., Zinkevich, M., Polyzotis, N., Whang, S., Roy, S.: Data validation for machine learning. In: Proceedings of SysML (2019), \url{https://mlsys.org/Conferences/2019/doc/2019/167.pdf}

\bibitem{bu_2018}
Bu, Y., Lu, J., Veeravalli, V.V.: Model change detection with application to machine learning (2018)

\bibitem{cohen_1995}
Cohen, W.W.: Fast effective rule induction. In: Prieditis, A., Russell, S. (eds.) Machine Learning Proceedings 1995, pp. 115--123. Morgan Kaufmann, San Francisco (CA) (1995)

\bibitem{friedman_2008}
Friedman, J.H., Popescu, B.E.: {Predictive learning via rule ensembles}. The Annals of Applied Statistics  \textbf{2}(3),  916 -- 954 (2008)

\bibitem{geng_2019}
Geng, J., Zhang, B., Huie, L.M., Lai, L.: Online change-point detection of linear regression models. IEEE Transactions on Signal Processing  \textbf{67}(12),  3316--3329 (2019). \doi{10.1109/TSP.2019.2914893}

\bibitem{harel_2014}
Harel, M., Crammer, K., El-Yaniv, R., Mannor, S.: Concept drift detection through resampling. In: Proceedings of the 31st International Conference on International Conference on Machine Learning - Volume 32. p. II–1009–II–1017. ICML'14, JMLR.org (2014)

\bibitem{harries_1999}
Harries, M., Nsw-cse tr, U., Wales, N.: Splice-2 comparative evaluation: Electricity pricing  (05 2003)

\bibitem{hinder_2023}
Hinder, F., Vaquet, V., Brinkrolf, J., Hammer, B.: Model based explanations of concept drift (2023)

\bibitem{kulesza_mental_2012}
Kulesza, T., Stumpf, S., Burnett, M., Kwan, I.: Tell me more? the effects of mental model soundness on personalizing an intelligent agent. In: Proceedings of the SIGCHI Conference on Human Factors in Computing Systems. p. 1–10. CHI '12, Association for Computing Machinery, New York, NY, USA (2012). \doi{10.1145/2207676.2207678}, \url{https://doi.org/10.1145/2207676.2207678}

\bibitem{lundberg_2017}
Lundberg, S., Lee, S.I.: A {Unified} {Approach} to {Interpreting} {Model} {Predictions} (2017), \url{http://arxiv.org/abs/1705.07874}, arXiv:1705.07874 [cs, stat]

\bibitem{marx_predictive_multiplicity_2020}
Marx, C., Calmon, F., Ustun, B.: Predictive multiplicity in classification. In: III, H.D., Singh, A. (eds.) Proceedings of the 37th International Conference on Machine Learning. Proceedings of Machine Learning Research, vol.~119, pp. 6765--6774. PMLR (13--18 Jul 2020), \url{https://proceedings.mlr.press/v119/marx20a.html}

\bibitem{miller_explanation_2018}
Miller, T.: Explanation in artificial intelligence: Insights from the social sciences. Artificial Intelligence  \textbf{267},  1--38 (2019). \doi{https://doi.org/10.1016/j.artint.2018.07.007}, \url{https://www.sciencedirect.com/science/article/pii/S0004370218305988}

\bibitem{muschalik_2022}
Muschalik, M., Fumagalli, F., Hammer, B., H{\"u}llermeier, E.: Agnostic {Explanation} of {Model} {Change} based on {Feature} {Importance}. KI - K{\"u}nstliche Intelligenz  \textbf{36}(3),  211--224 (Dec 2022)

\bibitem{leventi_2022}
Müller, S., Toborek, V., Beckh, K., Jakobs, M., Bauckhage, C., Welke, P.: An Empirical Evaluation of the Rashomon Effect in Explainable Machine Learning, pp. 462--478 (09 2023). \doi{10.1007/978-3-031-43418-1_28}

\bibitem{nair_2021}
Nair, R., Mattetti, M., Daly, E., Wei, D., Alkan, O., Zhang, Y.: What changed? interpretable model comparison. In: Zhou, Z.H. (ed.) Proceedings of the Thirtieth International Joint Conference on Artificial Intelligence, {IJCAI-21}. pp. 2855--2861. International Joint Conferences on Artificial Intelligence Organization (8 2021)

\bibitem{renard_understanding_2021}
Renard, X., Laugel, T., Detyniecki, M.: Understanding prediction discrepancies in machine learning classifiers (2021)

\bibitem{ribeiro_2016_lime}
Ribeiro, M.T., Singh, S., Guestrin, C.: "why should i trust you?": Explaining the predictions of any classifier. In: Proceedings of the 22nd ACM SIGKDD International Conference on Knowledge Discovery and Data Mining. p. 1135–1144. KDD '16, Association for Computing Machinery, New York, NY, USA (2016). \doi{10.1145/2939672.2939778}, \url{https://doi.org/10.1145/2939672.2939778}

\bibitem{ribiero_2018}
Ribeiro, M.T., Singh, S., Guestrin, C.: Anchors: High-precision model-agnostic explanations. Proceedings of the AAAI Conference on Artificial Intelligence  \textbf{32}(1) (Apr 2018)

\bibitem{wang_2022}
Wang, L., Wang, J., Zheng, Y., Jain, S., Yeh, C.C.M., Zhuang, Z., Ebrahimi, J., Zhang, W.: Learning from disagreement for event detection. In: 2022 IEEE International Conference on Big Data (Big Data). pp. 2411--2418 (2022). \doi{10.1109/BigData55660.2022.10020960}

\end{thebibliography}

\section{Supplementary Experimental Results}

This section provides the results obtained all other data sets.  For the synthetic AGRAWAL~\cite{agrawal_1993} data, we consider three different classification functions, where each function leads to a different distribution of the samples generated.

\begin{table}
\caption{Full Experiment results for Cover type (averaged over 10 runs)}\label{tab_result_cover_type}
\resizebox{\textwidth}{!}{%
\begin{tabular}{|l|l|l|l|l|l|l|l|}
\hline
Scenario & Min samples & Acc &Prec & Rec &\textbf{\#r} & Mean \textbf{\#l} & Mean \textbf{cov} \\

\hline
S1 &1 sample & \underline{\textbf{0.98}} $\pm$ 0.0 &\underline{\textbf{0.57}} $\pm$ 0.17 &\underline{\textbf{0.41}} $\pm$ 0.15 & 26.1 $\pm$ 10.28 &7.15 $\pm$ 1.26 &0.07\% $\pm$ 0.03 \\
S1 & 0.1 \% &0.98 $\pm$ 0.01 &0.52 $\pm$ 0.17 & 0.4 $\pm$ 0.19 &12.1 $\pm$ 6.21 &6.11 $\pm$ 1.22 &0.14\% $\pm$ 0.03 \\
S1 & 1 \% &0.19 $\pm$ 0.41 &0.12 $\pm$ 0.27 &0.08 $\pm$ 0.19 & 0.4 $\pm$ 0.97 & 0.6 $\pm$ 1.26 &0.23\% $\pm$ 0.49 \\
S1 & 2.5 \% &0.1 $\pm$ 0.3 &0.05 $\pm$ 0.16 & 0.03 $\pm$ 0.1 & 0.1 $\pm$ 0.32 & 0.2 $\pm$ 0.63 &\underline{\textbf{0.28\%}} $\pm$ 0.87 \\
S1 &10 \% &- &- &- &0.0 &-&- \\
S1 &25 \% &- &- &- &0.0 &-&- \\
\hline
S2 &1 sample &0.68 $\pm$ 0.06 & 0.6 $\pm$ 0.05 &0.56 $\pm$ 0.04 &401.8 $\pm$ 90.57 &9.74 $\pm$ 0.51 &0.09\% $\pm$ 7e-03 \\
S2 & 0.1 \% &0.69 $\pm$ 0.07 & 0.6 $\pm$ 0.03 &\underline{\textbf{0.58}} $\pm$ 0.04 &227.6 $\pm$ 68.64 &8.73 $\pm$ 0.65 &0.17\% $\pm$ 0.01 \\
S2 & 1 \% & \underline{\textbf{0.71}} $\pm$ 0.1 &\underline{\textbf{0.68}} $\pm$ 0.08 &0.54 $\pm$ 0.09 & 25.2 $\pm$ 10.45 &5.82 $\pm$ 0.91 &1.318\% $\pm$ 0.064 \\
S2 & 2.5 \% & 0.7 $\pm$ 0.09 &0.66 $\pm$ 0.09 &0.52 $\pm$ 0.08 &10.0 $\pm$ 3.56 &4.54 $\pm$ 0.97 &3.168\% $\pm$ 0.211 \\
S2 &10 \% & 0.68 $\pm$ 0.1 &0.64 $\pm$ 0.09 &0.46 $\pm$ 0.11 & 2.3 $\pm$ 0.95 &2.42 $\pm$ 0.92 & 12.79\% $\pm$ 1.939 \\
S2 &25 \% &0.52 $\pm$ 0.29 & 0.5 $\pm$ 0.28 &0.38 $\pm$ 0.22 & \underline{\textbf{0.8}}$\pm$ 0.42 & \underline{\textbf{1.0}}$\pm$ 0.67 & \underline{\textbf{26.1\%}} $\pm$14.65 \\
\hline
S3 &1 sample & \underline{\textbf{0.99}} $\pm$ 0.0 &\underline{\textbf{0.97}} $\pm$ 0.02 &\underline{\textbf{0.95}} $\pm$ 0.04 & 15.2 $\pm$ 5.9 & 5.22 $\pm$ 0.8 &0.83\% $\pm$ 0.47 \\
S3 & 0.1 \% & 0.99 $\pm$ 0.0 &0.96 $\pm$ 0.03 &0.95 $\pm$ 0.03 &13.6 $\pm$ 5.04 &4.98 $\pm$ 0.75 &0.95\% $\pm$ 0.63 \\
S3 & 1 \% &0.98 $\pm$ 0.01 &0.89 $\pm$ 0.07 &0.91 $\pm$ 0.06 & 5.6 $\pm$ 1.51 & 3.8 $\pm$ 0.47 &2\% $\pm$ 0.7 \\
S3 & 2.5 \% &0.97 $\pm$ 0.01 & 0.87 $\pm$ 0.1 &0.85 $\pm$ 0.13 & 2.8 $\pm$ 1.03 &2.87 $\pm$ 0.52 &3.84\% $\pm$ 0.62 \\
S3 &10 \% & 0.75 $\pm$ 0.4 &0.56 $\pm$ 0.32 &0.77 $\pm$ 0.41 & 0.8 $\pm$ 0.42 & 0.8 $\pm$ 0.42 &\underline{\textbf{13.59\%}} $\pm$ 7.29 \\
S3 &25 \% &0.27 $\pm$ 0.43 &0.17 $\pm$ 0.27 & 0.3 $\pm$ 0.48 & \underline{\textbf{0.3}}$\pm$ 0.48 & \underline{\textbf{0.3}}$\pm$ 0.48 &7.46\% $\pm$ 12.009 \\
\hline

\end{tabular}%
}
\end{table}

\begin{table}
\caption{Full Experiment results for Agrawal using the classification function 0 (averaged over 10 runs)}\label{tab_result_agrawal_0}
\resizebox{\textwidth}{!}{%
\begin{tabular}{|l|l|l|l|l|l|l|l|}
\hline
Scenario & Min samples & Acc & Prec & Rec & +\textbf{\#r} & Mean \textbf{\#l} & Mean +\textbf{cov} \\ \hline
S1 & 1 sample & \underline{\textbf{0.99}}$\pm$0.0& 0.21$\pm$0.05 & \underline{\textbf{0.12}}$\pm$0.06& 14.5$\pm$3.69 & 6.37$\pm$1.05 & 0.04\%$\pm$7e-03 \\
S1 & 0.1\% & \underline{\textbf{0.99}}$\pm$0.0& \underline{\textbf{0.28}}$\pm$0.21& 0.11$\pm$0.1 & 3.2$\pm$1.55 & 5.26$\pm$1.17 & \underline{\textbf{0.11\%}}$\pm$0.01 \\
S1 & 1\% & - & - & - & 0.0& -& - \\
S1 & 2.5\% & - & - & - &0.0&-& - \\
S1 & 10\% & - & - & - &0.0&-& - \\
S1 & 25\% & - & - & - &0.0&-& - \\
\hline
S2 & 1 sample & 0.68$\pm$0.01 & 0.61$\pm$0.01 & 0.56$\pm$0.01 & 452.8$\pm$16.44 & 8.74$\pm$0.36 & 0.08\%$\pm$2e-03 \\
S2 & 0.1\% & 0.7$\pm$0.01 & 0.64$\pm$0.02 & 0.6$\pm$0.01 & 236.9$\pm$8.72 & 8.32$\pm$0.46 & 0.16\%$\pm$4e-03 \\
S2 & 1\% & 0.78$\pm$0.0 & 0.78$\pm$0.01 & \underline{\textbf{0.64}}$\pm$0.01& 25.0$\pm$1.15 & 6.37$\pm$0.48 & 1.32\%$\pm$0.06 \\
S2 & 2.5\% & \underline{\textbf{0.79}}$\pm$0.01& \underline{\textbf{0.79}}$\pm$0.01& \underline{\textbf{0.64}}$\pm$0.01& 10.1$\pm$0.57 & 5.26$\pm$0.46 & 3.24\%$\pm$0.17 \\
S2 & 10\% & \underline{\textbf{0.79}}$\pm$0.01& \underline{\textbf{0.79}}$\pm$0.01& \underline{\textbf{0.64}}$\pm$0.01& 2.7$\pm$0.48 & 3.2$\pm$0.53 & 12.52\%$\pm$2.62 \\
S2 & 25\% & \underline{\textbf{0.79}}$\pm$0.01& \underline{\textbf{0.79}}$\pm$0.01& \underline{\textbf{0.64}}$\pm$0.01& \underline{\textbf{1.0}}$\pm$0.0& \underline{\textbf{2.0}}$\pm$0.0& \underline{\textbf{32.66\%}}$\pm$0.6 \\
\hline
S3 & 1 sample & \underline{\textbf{1.0}}$\pm$0.0& \underline{\textbf{1.0}}$\pm$0.0& \underline{\textbf{1.0}}$\pm$0.0& 1.0$\pm$0.0 & 2.0$\pm$0.0 & \underline{\textbf{6.42\%}}$\pm$0.24\\
S3 & 0.1\% & \underline{\textbf{1.0}}$\pm$0.0& \underline{\textbf{1.0}}$\pm$0.0& \underline{\textbf{1.0}}$\pm$0.0& 1.0$\pm$0.0 & 2.0$\pm$0.0 & \underline{\textbf{6.42\%}}$\pm$0.24\\
S3 & 1\% & \underline{\textbf{1.0}}$\pm$0.0& \underline{\textbf{1.0}}$\pm$0.0& \underline{\textbf{1.0}}$\pm$0.0& 1.0$\pm$0.0 & 2.0$\pm$0.0 & \underline{\textbf{6.42\%}}$\pm$0.24\\
S3 & 2.5\% & \underline{\textbf{1.0}}$\pm$0.0& \underline{\textbf{1.0}}$\pm$0.0& \underline{\textbf{1.0}}$\pm$0.0& 1.0$\pm$0.0 & 2.0$\pm$0.0 & \underline{\textbf{6.42\%}}$\pm$0.24\\
S3 & 10\% & 0.09$\pm$0.29 & 0.05$\pm$0.15 & 0.1$\pm$0.32 & 0.1$\pm$0.32 & 0.1$\pm$0.32 & 1.31\%$\pm$4.16 \\
S3 & 25\% & - & - & - & 0.0&0.0& - \\ 
\hline
\end{tabular}%
}
\end{table}

\begin{table}
\caption{Full Experiment results for Agrawal using the classification function 1 (averaged over 10 runs)}\label{tab_result_agrawal_1}
\resizebox{\textwidth}{!}{%
\begin{tabular}{|l|l|l|l|l|l|l|l|}
\hline
Scenario & Min samples & Acc &Prec & Rec &\textbf{\#r} & Mean \textbf{\#l} & Mean \textbf{cov} \\

\hline
S1 &1 sample & 0.98 $\pm$ 0.0 &0.11 $\pm$ 0.07 &\underline{\textbf{0.07}} $\pm$ 0.04 & 17.2 $\pm$ 2.9 &7.13 $\pm$ 1.01 &0.04\% $\pm$ 4e-03 \\
S1 & 0.1 \% & \underline{\textbf{0.99}} $\pm$ 0.0 &\underline{\textbf{0.19}} $\pm$ 0.21 &0.04 $\pm$ 0.04 & 2.7 $\pm$ 0.67 &6.42 $\pm$ 2.08 &\underline{\textbf{0.11\%}} $\pm$ 0.02 \\
S1 & 1 \% &- &- &- &0.0 &- &- \\
S1 & 2.5 \% &- &- &- &0.0 &- &- \\
S1 &10 \% &- &- &- &0.0 &- &- \\
S1 &25 \% &- &- &- &0.0 &- &- \\
\hline
S2 &1 sample &0.62 $\pm$ 0.01 &0.59 $\pm$ 0.02 &0.53 $\pm$ 0.02 &542.0 $\pm$ 12.23 &8.29 $\pm$ 0.18 &0.08\% $\pm$ 2e-03 \\
S2 & 0.1 \% &0.63 $\pm$ 0.01 & 0.6 $\pm$ 0.01 &0.57 $\pm$ 0.02 & 296.6 $\pm$ 8.37 &7.71 $\pm$ 0.27 &0.15\% $\pm$ 2e-03 \\
S2 & 1 \% &\underline{\textbf{0.68}} $\pm$ 0.02 &\underline{\textbf{0.67}} $\pm$ 0.03 &\underline{\textbf{0.59}} $\pm$ 0.03 &30.6 $\pm$ 1.58 &5.59 $\pm$ 0.25 &1.32\% $\pm$ 0.043 \\
S2 & 2.5 \% &0.67 $\pm$ 0.03 &\underline{\textbf{0.67}} $\pm$ 0.05 &0.58 $\pm$ 0.03 &12.5 $\pm$ 1.27 & 4.6 $\pm$ 0.27 & 3.26\% $\pm$ 0.285 \\
S2 &10 \% & 0.6 $\pm$ 0.02 &0.59 $\pm$ 0.02 &0.43 $\pm$ 0.07 & 2.8 $\pm$ 0.63 &3.47 $\pm$ 0.38 & 12.11\% $\pm$ 1.52 \\
S2 &25 \% &0.59 $\pm$ 0.01 &0.56 $\pm$ 0.02 &0.53 $\pm$ 0.06 &\underline{\textbf{1.0 }}$\pm$ 0.0 &\underline{\textbf{2.0 }}$\pm$ 0.0 &\underline{\textbf{43.61\%}} $\pm$ 5.92 \\
\hline
S3 &1 sample & \underline{\textbf{1.0 }}$\pm$ 0.01 &\underline{\textbf{0.98}} $\pm$ 0.03 &\underline{\textbf{0.97}} $\pm$ 0.03 &10.5 $\pm$ 6.77 & 4.52 $\pm$ 1.3 & 0.96\% $\pm$ 0.43 \\
S3 & 0.1 \% &0.99 $\pm$ 0.01 &0.97 $\pm$ 0.04 &\underline{\textbf{0.97}} $\pm$ 0.04 & 9.8 $\pm$ 5.31 & 4.5 $\pm$ 1.15 &0.95\% $\pm$ 0.37 \\
S3 & 1 \% &0.98 $\pm$ 0.01 & 0.9 $\pm$ 0.06 &0.88 $\pm$ 0.09 & 5.6 $\pm$ 0.84 & 3.6 $\pm$ 0.19 &1.34\% $\pm$ 0.16 \\
S3 & 2.5 \% & 0.95 $\pm$ 0.0 &0.72 $\pm$ 0.09 &0.54 $\pm$ 0.09 & 2.1 $\pm$ 0.57 &3.37 $\pm$ 0.35 &\underline{\textbf{2.8\%}} $\pm$ 0.28 \\
S3 &10 \% &- &- &- &0.0 &- &- \\
S3 &25 \% &- &- &- &0.0 &- &- \\
\hline

\end{tabular}%
}
\end{table}

\begin{table}
\caption{Full Experiment results for Agrawal using the classification function 2 (averaged over 10 runs)}\label{tab_result_agrawal_2}
\resizebox{\textwidth}{!}{%
\begin{tabular}{|l|l|l|l|l|l|l|l|}
\hline
Scenario & Min samples & Acc &Prec & Rec &\textbf{\#r} & Mean \textbf{\#l} & Mean \textbf{cov} \\
\hline

S1 &1 sample & \underline{\textbf{0.99}} $\pm$ 0.0 &0.05 $\pm$ 0.16 &0.01 $\pm$ 0.02 & 3.9 $\pm$ 2.56 &5.33 $\pm$ 0.96 &0.04\% $\pm$ 0.02 \\
S1 & 0.1 \% & 0.5 $\pm$ 0.52 & \underline{\textbf{0.1}}$\pm$ 0.22 &\underline{\textbf{0.02}} $\pm$ 0.05 &0.6 $\pm$ 0.7 & 1.7 $\pm$ 1.95 &\underline{\textbf{0.05\%}} $\pm$ 0.06 \\
S1 & 1 \% &- &- &- &0.0 &-&- \\
S1 & 2.5 \% &- &- &- &0.0 &-&- \\
S1 &10 \% &- &- &- &0.0 &-&- \\
S1 &25 \% &- &- &- &0.0 &-&- \\
\hline
S2 &1 sample &0.55 $\pm$ 0.02 &0.55 $\pm$ 0.02 & 0.5 $\pm$ 0.03 &631.7 $\pm$ 20.85 &9.09 $\pm$ 0.23 &0.07\% $\pm$ 2e-03 \\
S2 & 0.1 \% &0.55 $\pm$ 0.02 &0.54 $\pm$ 0.02 &0.53 $\pm$ 0.04 &355.9 $\pm$ 14.54 &8.42 $\pm$ 0.29 &0.14\% $\pm$ 3e-03 \\
S2 & 1 \% &0.55 $\pm$ 0.02 &0.55 $\pm$ 0.02 &0.53 $\pm$ 0.04 &36.2 $\pm$ 2.62 &5.93 $\pm$ 0.55 &13.12\% $\pm$ 0.05 \\
S2 & 2.5 \% &\underline{\textbf{0.56}} $\pm$ 0.03 &\underline{\textbf{0.56}} $\pm$ 0.03 &0.55 $\pm$ 0.07 & 14.8 $\pm$ 2.1 &4.63 $\pm$ 0.59 &3.37\% $\pm$ 0.22 \\
S2 &10 \% &0.55 $\pm$ 0.02 &0.54 $\pm$ 0.02 &\underline{\textbf{0.58}} $\pm$ 0.07 & 3.8 $\pm$ 0.92 &2.41 $\pm$ 0.44 &14.18\% $\pm$ 1.39 \\
S2 &25 \% &0.47 $\pm$ 0.17 &0.47 $\pm$ 0.17 &0.49 $\pm$ 0.25 & \underline{\textbf{1.5}}$\pm$ 0.71 & \underline{\textbf{1.4}}$\pm$ 0.57 &\underline{\textbf{27.9\%}} $\pm$ 10.47 \\
\hline
S3 &1 sample &\underline{\textbf{1.0}}$\pm$ 0.0 &\underline{\textbf{1.0}}$\pm$ 0.0 &\underline{\textbf{1.0}}$\pm$ 0.0 &4.4 $\pm$ 0.7 &3.53 $\pm$ 0.22 &2.16\% $\pm$ 0.29 \\
S3 & 0.1 \% &1.0 $\pm$ 0.0 &\underline{\textbf{1.0}}$\pm$ 0.0 &\underline{\textbf{1.0}}$\pm$ 0.0 &4.4 $\pm$ 0.7 &3.54 $\pm$ 0.24 &2.16\% $\pm$ 0.29 \\
S3 & 1 \% &\underline{\textbf{1.0}}$\pm$ 0.0 &0.99 $\pm$ 0.01 &0.99 $\pm$ 0.01 & 4.3 $\pm$ 0.48 &3.46 $\pm$ 0.25 &2.19\% $\pm$ 0.23 \\
S3 & 2.5 \% &0.95 $\pm$ 0.01 &0.71 $\pm$ 0.08 &0.81 $\pm$ 0.12 & 3.3 $\pm$ 0.82 &2.97 $\pm$ 0.23 &\underline{\textbf{3.43\%}} $\pm$ 0.64 \\
S3 &10 \% &0.18 $\pm$ 0.38 &0.11 $\pm$ 0.23 &0.11 $\pm$ 0.23 & 0.2 $\pm$ 0.42 & 0.4 $\pm$ 0.84 &1.99 $\pm$ 4.19 \\
S3 &25 \% &- &- &- &0.0 &-&- \\
\hline

\end{tabular}%
}
\end{table}

\end{document}